\documentclass[sigconf]{acmart}

\renewcommand\footnotetextcopyrightpermission[1]{}
\usepackage{wrapfig}
\usepackage{booktabs}       
\usepackage{amsfonts}       
\usepackage{nicefrac}       
\usepackage{amsmath}

\usepackage{amssymb}
\usepackage{array}
\usepackage{multirow}
\usepackage{color}
\usepackage{colortbl}
\usepackage{framed}
\usepackage{bm}
\usepackage{xspace}
\usepackage{enumitem}
\usepackage{xparse}
\usepackage{tcolorbox}
\usepackage[english]{babel}        
\usepackage{microtype}             
\usepackage{hyphenat}              
\usepackage{url}
\usepackage{pifont}
\usepackage{mathtools}
\usepackage{lipsum}
\usepackage{adjustbox}
\usepackage{stfloats}
\usepackage{algorithm}
\usepackage{listings} 
\usepackage{graphicx} 
\usepackage{subcaption} 
\usepackage{tikz} 
\usetikzlibrary{positioning}
\usepackage{algpseudocode}
\newcommand{\cmark}{\ding{51}}%
\newcommand{\xmark}{\ding{55}}%

\definecolor{Gray}{gray}{0.2}
\definecolor{lightgray}{gray}{0.92}
\definecolor{OurColor}{rgb}{0.886, 0.941, 0.851}

\definecolor{customgray}{gray}{0.35}
\newcommand{\KB}[1]{{\color{customgray}#1}}

\newcommand{\tit}[1]{\smallbreak\noindent\textbf{#1.}}

\newcommand{\ours}{MMKB-RAG\xspace}

\definecolor{backcolour}{rgb}{0.96,0.96,0.96}
\definecolor{GrayL}{gray}{0.3}
\definecolor{coolblack}{rgb}{0.0, 0.18, 0.39}
\definecolor{cornellred}{rgb}{0.7, 0.11, 0.11}
\definecolor{darktangerine}{rgb}{1.0, 0.66, 0.07}
\definecolor{deepcarrotorange}{rgb}{0.91, 0.41, 0.17}
\definecolor{noretcolor}{rgb}{0.847, 0.431, 0.8}
\definecolor{retcolor}{rgb}{0.627, 0.7686, 1.0}
\definecolor{relcolor}{RGB}{0, 176, 80}
\definecolor{imagecolor}{RGB}{254, 202, 203}
\definecolor{passagecolor}{RGB}{187, 232, 172}

\lstdefinestyle{mystyle}{
    backgroundcolor=\color{backcolour},   
    basicstyle=\ttfamily\footnotesize,
    breakatwhitespace=false,         
    breaklines=true,                 
    captionpos=b,                    
    keepspaces=true,                                 
    showspaces=false,                
    showstringspaces=false,
    showtabs=false,                  
    tabsize=2,
    escapeinside={(*}{*)},
}
\AtBeginDocument{%
  }

\copyrightyear{}
\acmYear{}
\acmDOI{}
\acmConference[]{}{}{}
\acmISBN{}

\begin{document}

\title{MMKB-RAG: A Multi-Modal Knowledge-Based Retrieval-Augmented Generation Framework}


\author{Zihan Ling}
\affiliation{
  \institution{Peking University}
  \country{China}
  }
\email{lingzihan@stu.pku.edu.cn}

\author{Zhiyao Guo$^\dagger$}
\affiliation{%
  \institution{Alibaba Group}
  \country{China}
  }
\email{guozhiyao45@gmail.com}

\author{Yixuan Huang}
\affiliation{%
  \institution{Alibaba Group}
  \country{China}
}
\email{huangyixuan@sjtu.edu.cn}

\author{Yi An}
\affiliation{%
  \institution{Peking University}
  \country{China}
  }
\email{anyi@stu.pku.edu.cn}

\author{Shuai Xiao}
\affiliation{%
  \institution{Alibaba Group}
  \country{China}
}
\email{shuai.xsh@gmail.com}

\author{Jinsong Lan}
\affiliation{%
  \institution{Alibaba Group}
  \country{China}
}
\email{jinsonglan.ljs@taobao.com}

\author{Xiaoyong Zhu}
\affiliation{%
  \institution{Alibaba Group}
  \country{China}
}
\email{xiaoyong.z@taobao.com}

\author{Bo Zheng$^\ast$}
\affiliation{%
  \institution{Alibaba Group}
  \country{China}
}
\email{bozheng@alibaba-inc.com}


\begin{abstract}
Recent advancements in large language models (LLMs) and multi-modal LLMs have been remarkable. However, these models still rely solely on their parametric knowledge, which limits their ability to generate up-to-date information and increases the risk of producing erroneous content. Retrieval-Augmented Generation (RAG) partially mitigates these challenges by incorporating external data sources, yet the reliance on databases and retrieval systems can introduce irrelevant or inaccurate documents, ultimately undermining both performance and reasoning quality. In this paper, we propose Multi-Modal Knowledge-Based Retrieval-Augmented Generation (MMKB-RAG), a novel multi-modal RAG framework that leverages the inherent knowledge boundaries of models to dynamically generate semantic tags for the retrieval process. This strategy enables the joint filtering of retrieved documents, retaining only the most relevant and accurate references. Extensive experiments on knowledge-based visual question-answering tasks demonstrate the efficacy of our approach: on the E-VQA dataset, our method improves performance by +4.2\% on the Single-Hop subset and +0.4\% on the full dataset, while on the InfoSeek dataset, it achieves gains of +7.8\% on the Unseen-Q subset, +8.2\% on the Unseen-E subset, and +8.1\% on the full dataset. These results highlight significant enhancements in both accuracy and robustness over the current state-of-the-art MLLM and RAG frameworks.
\end{abstract}








\maketitle

\section{Introduction}

\begin{figure}[t]
  \centering
  \includegraphics[width=0.44\textwidth]{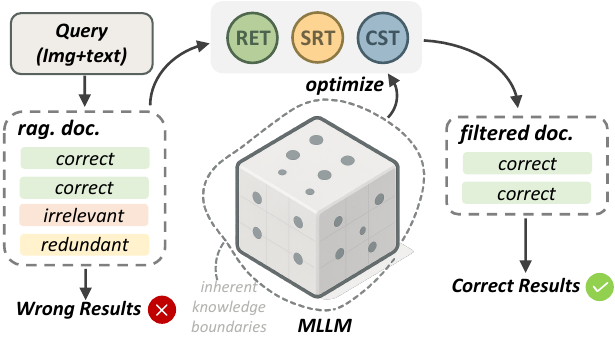}
      \vspace{-.35cm}
  \caption{Examples illustrate that RAG may retrieve irrelevant and redundant documents, resulting in incorrect outcomes. Our approach leverages the MLLM's knowledge boundaries to filter references, retaining essential evidence for accurate answers.}
  \label{fig:teaser}
      \vspace{-.6cm}
\end{figure}

\footnotetext[1]{$\ast$ Corresponding author}
\footnotetext[2]{$\dagger$ Project leader}

The rapid evolution of large language models (LLMs) and multi-modal large language models (MLLMs) has revolutionized both natural language processing and visual reasoning tasks. Trained on extensive datasets, these models excel at leveraging intrinsic parametric knowledge to generate coherent and contextually appropriate responses. However, a fundamental challenge remains: The reliance on static parametric knowledge  (i.e., the knowledge acquired during pre-training) often leads to errors or hallucinations \cite{rawte-etal-2023-troubling}, particularly when addressing complex queries that demand precise, domain-specific, or real-time information\cite{li2025benchmarkingmultimodalretrievalaugmented}. For example, in knowledge-based visual question-answering (VQA) tasks such as Encyclopedic-VQA (E-VQA) \citep{mensink2023encyclopedicvqavisualquestions} and InfoSeek \citep{chen2023pretrainedvisionlanguagemodels}, even state-of-the-art MLLMs struggle to generate accurate responses because of lacking the ability to retrieve and integrate external knowledge effectively. This deficiency underscores a critical gap in current MLLM architectures: the need for mechanisms that enable dynamic and reliable access to external knowledge sources.

To overcome the limitations of static parametric knowledge, researchers have developed Retrieval-Augmented Generation (RAG) frameworks that enables models to incorporate the latest and most relevant information during inference. These frameworks can be broadly divided into retrieval-side mechanisms and generation-side mechanisms. 
On the retrieval side, existing methods primarily focus on aligning visual and textual modalities with external knowledge sources. CLIP-based architectures leverage contrastive image-text encoders to establish coarse-grained alignments between image-question pairs and knowledge entries, while Dense Passage Retrieval (DPR)-based architectures \cite{lin2023fine,lin2024preflmr} enhance precision by incorporating fine-grained features. For instance, Wiki-LLaVA \cite{caffagni2024wiki} employs a CLIP-based multi-stage retrieval pipeline to improve alignment, and RoRA-VLM \cite{qi2024rora} utilizes adversarial training techniques to enhance robustness against irrelevant content. Despite these advancements, achieving fine-grained alignment between visual and textual information remains challenging, often leading to suboptimal retrieval results.

On the generation side, researchers have sought to improve RAG systems by enabling MLLMs to evaluate the relevance and accuracy of the retrieved content autonomously. For example, ReflectiVA \cite{cocchi2024augmenting} uses specialized tokens to steer the retrieval process, while EchoSight \cite{Yan_2024} incorporates fine-tuned re-ranking modules to filter out noisy or irrelevant documents. However, these self-evaluation strategies typically depend on external annotation pipelines or auxiliary models, which may not fully capture the intrinsic knowledge boundaries of the MLLM. This observation highlights the demand for an integrated framework that can use the inherent knowledge of MLLMs to guide both retrieval and filtering processes dynamically.

Towards the problems above, we propose a novel framework named Multi-Modal Knowledge-Based Retrieval-Augmented Generation (MMKB-RAG). Unlike traditional RAG systems that rely solely on external retrieval strategies, MMKB-RAG leverages the inherent knowledge boundaries of the target MLLM to dynamically generate specialized tags for filtering retrieved documents. Through its endogenous annotation system, the MLLM autonomously determines when retrieval is necessary and verifies both the consistency and relevance of the retrieved knowledge, all based on its own intrinsic knowledge limits. The key innovation of MMKB-RAG is its ability to bridge the gap between parametric and retrieved knowledge by transitioning from exogenous, auxiliary-model-dependent annotation pipelines to an intrinsic, capability-aware system, addressing the shortcomings of both multi-modal retrieval and self-evaluation approaches.


Overall, this paper presents the following main contributions:
\begin{itemize}[noitemsep, topsep=0pt]
    \item \textbf{Token System Framework}: MMKB-RAG introduces a three-stage process to determine retrieval necessity, evaluate the relevance of individual documents, and verify the consistency among multiple documents, ensuring accurate and robust reasoning.
    \item \textbf{Internal Knowledge Utilization}: By leveraging the inherent knowledge of MLLMs and their interactions with datasets, MMKB-RAG autonomously defines knowledge boundaries and guides retrieval without relying on external annotations.
    \item \textbf{Superior Performance}: MMKB-RAG outperforms state-of-the-art models in knowledge-based VQA tasks, particularly in handling fine-grained factual queries and complex multi-modal reasoning challenges.
\end{itemize}

\section{Related Work}

\tit{Multi-modal LLM}
The emergence of large language models (LLMs)\-\citep{dubey2024llama,jiang2023mistral,yang2024qwen2} has catalyzed significant progress in multi-modal LLMs (MLLMs)\citep{liu2024visual,wang2024qwen2,chen2024internvl}, enabling basic visual comprehension and commonsense reasoning capabilities. Notably, implementations such as LLaVA \citep{liu2024visual}, Qwen-VL\citep{wang2024qwen2}, and InternVL\citep{chen2024internvl} have demonstrated strong performance on standard visual question answering (VQA) benchmarks.

\tit{Knowledge-based VQA}
Knowledge-based VQA tasks require MLLMs to integrate information beyond visual content by leveraging external knowledge sources. Early benchmarks such as KVQA\citep{shah2019kvqa}, OK-VQA\citep{marino2019ok}, and A-OKVQA\citep{schwenk2022okvqa} focused primarily on commonsense reasoning, an area where large-scale pre-trained MLLMs perform effectively thanks to their implicit knowledge representations. Recent datasets like E-VQA \citep{mensink2023encyclopedicvqavisualquestions} and InfoSeek \citep{chen2023pretrainedvisionlanguagemodels} have pushed the field toward Wikipedia-scale knowledge integration, necessitating a comprehensive understanding of specific Wikipedia entities and fine-grained details.  Although these benchmarks highlight the upper bounds of current systems, even state-of-the-art MLLMs remain constrained by their parameter-intensive architectures when processing detailed factual queries. This fundamental limitation frequently results in the generation of hallucinated content\citep{liu2024survey,rawte2023survey,tong2024eyes}. 

The RAG framework\citep{lewis2020retrieval} addresses these limitations by dynamically integrating external knowledge during inference. By employing multi-stage inference pipelines that merge parametric knowledge with retrieved information, RAG architectures exhibit considerable promise in mitigating these intrinsic constraints.

\tit{Advanced RAG}
Within the RAG paradigm, multi-modal retrieval mechanisms typically employ two principal strategies: 1) CLIP-based architectures employ contrastive image-text encoders\citep{sun2023eva} to establish coarse-grained alignment between image-question pairs and knowledge entries, and 2) DPR-based architectures\citep{lin2023fine,lin2024preflmr} incorporate fine-grained visual features (e.g., Regions-of-Interest) to enhance retrieval precision. The retrieved entries are subsequently integrated into MLLM as contextual references.

Recently, Wiki-LLaVA\citep{caffagni2024wiki} has integrated knowledge through CLIP-based multi-stage retrieval process, while RoRA-VLM\citep{qi2024rora} employs adversarial training to improve robustness against irrelevant retrieved content via query-oriented visual token pruning. Meanwhile, EchoSight\citep{Yan_2024} introduces multi-modal re-ranking modules through fine-tuned Q-Former. Although existing approaches primarily focus on optimizing multi-modal retrieval, they neglect the intrinsic role of MLLMs in self-evaluating knowledge boundaries. 

Inspired by self-RAG\citep{asai2023self}, we propose the MMKB-RAG framework where the MLLM autonomously generates specialized tokens to determine whether retrieval is necessary and verify both the consistency and relevance of the retrieved knowledge. Although the most recent work ReflectiVA\citep{cocchi2024augmenting} implements a similar specialized-token mechanism, its annotation pipeline fundamentally differs by leveraging different MLLMs, thereby neglecting the knowledge limitations of the target model. In contrast, our framework establishes an endogenous, knowledge boundary-aware annotation system that directly aligns label generation with the target model. This paradigm shift from exogenous, model-dependent annotation process to intrinsic, capability-aware labeling approach, yields significant performance gains.

\section{Proposed Method}

\begin{figure*}[t]
    \centering
    \includegraphics[width=\textwidth]{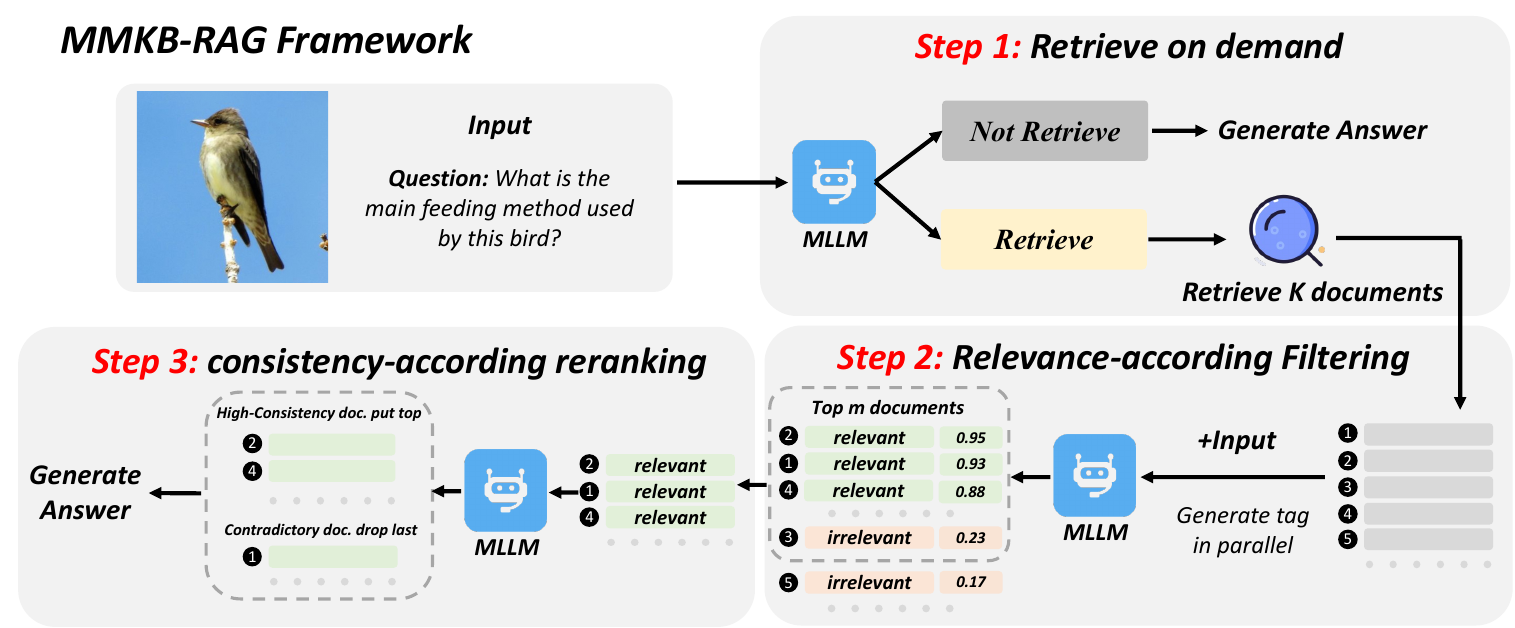}
    \vspace{-.6cm}
    \caption{MMKB-RAG Pipeline. Given an input query, MMKB-RAG initially assesses retrieval necessity. For retrieval-dependent queries, we employ a two-stage filtering process (Steps 2 and 3) to ensure only high-quality reference documents are preserved. This curated context is then provided to the MLLM to generate comprehensive and accurate responses.}
    \label{fig:framework}
    \vspace{-.35cm}
\end{figure*}

In Knowledge-based Visual Question Answering (VQA), the system processes an image-question pair $(I, Q)$ and leverages external knowledge to generate accurate responses. Traditional Multi-Modal Retrieval-Augmented Generation (MMRAG) approaches retrieve relevant documents from knowledge bases to support answer generation, formulated as:
\begin{equation}
    y_{\text{ans}} = \underset{y}{\arg\max}\ \text{MLLM}(y|I,Q,D_k,P_{vqa}).
    \label{eq:retrieval_augmented_generation}
\end{equation}
where $D_k=\{d_1, d_2,\ldots,d_{k}\}$ denotes the set of top-$k$ documents retrieved by retriever $R$ given the image-question pair $(I, Q)$, and $P_{vqa}$ represents the prompt template designed for visual question answering.

As illustrated in Fig.\ref{fig:framework}, our proposed MMKB-RAG (Multi-Modal Knowledge-Based Retrieval-Augmented Generation) extends beyond conventional MMRAG frameworks through three key innovations:  (1) leveraging the model's internal knowledge to determine when external retrieval is necessary, improving efficiency; (2) dynamically re-ranking retrieved documents based on their relevance to the input query; and (3) employing consistency checking to filter out inconsistent information, thereby improving answer reliability and accuracy. In the following sections, we detail our approach: Section 3.1 introduces the token system employed in MMKB-RAG, while Section 3.2 describes the model's training methodology.

\subsection{Token System for MMKB-RAG}
\tit{Retrieval Token (RET)}
Given an input pair $(I, Q)$, where $I$ represents the image and $Q$ represents the question text, we classify the input into two categories: (1) cases where the MLLM can answer using internal knowledge, represented by $[No Ret]$, and (2) cases that require external knowledge retrieval for an accurate response, represented by $[Ret]$. For inputs labeled as $[Ret]$, the retrieval $R$ fetches the top-$N$ most relevant documents, allowing the MLLM to incorporate external knowledge for a more accurate answer; for $[No Ret]$ inputs, the MLLM directly generates the answer, thereby optimizing computational efficiency. A fine-tuned MLLM predicts these tags with the designated prompt $P_{RET}$.
\begin{equation}
    RET = \text{MLLM}{_{ft}}(I, Q,P_{RET})
\end{equation}
we extract the logits corresponding to $[Ret]$ and $[NoRet]$ from the first generated token and apply a softmax function to convert them into normalized probability scores. The hyperparameter $\gamma$ is then used to modulate the final predicted class. By varying $\gamma$, we can control the model's propensity to perform retrieval operations. Typically, $\gamma$ is set to $0.5$.

\begin{equation}
    RET =
    \begin{cases}
        \text{$[Ret]$}, & \text{if } Score_{RET} > \gamma\\
        \text{$[NoRet]$}, & \text{otherwise}
    \end{cases}
\end{equation}

\begin{equation}
    Score_{RET} = \frac{\exp(z_{Ret})}{\exp(z_{Ret}) + \exp(z_{NoRet})}
\end{equation}


\tit{Single-Relevant Token Rerank (SRT)}
For inputs classified as $[Ret]$, the initially retrieved documents $D_k$ are selected based solely on embedding similarity scores. However, this may not align optimally with the internal knowledge of MLLM, potentially introducing irrelevant or contradictory information. We propose leveraging the MLLM's internal knowledge capabilities for relevance determination. To this end, we introduce $[Rel]$ and $[NoRel]$ tags, which enable the MLLM to evaluate each candidate document via a dedicated prompt $P_{SRT}$. For a given pair (I, Q), the prediction for the i-th document $d_i$ is formulated as follows:
\begin{equation}
    SRT^{i} = \text{MLLM}{_{ft}}(I, Q,d_{i},P_{SRT})
\end{equation}
Similarly to the previous section, we compute the softmax probabilities for the first-token logits of document $d_i$, where $z_{Rel}$ corresponds to $[Rel]$ and $z_{No Rel}$ to $[NoRel]$. We re-rank all $k$ retrieved documents based on the probabilities of $[Rel]$ and select the top-$K_{SRT}$ documents as our final result, where $k_{SRT} \leq k$. This process can be formulated as:
\begin{equation}
    D_{k_{SRT}} = \{d_i | i \in \arg\text{Top}k_{SRT}\{\text{Score}^{j}_{SRT}|j\in[1,k]\}\}
\end{equation}
\begin{equation}
    Score^{i}_{SRT} = \frac{\exp(z^{i}_{Rel})}{\exp(z^{i}_{Rel}) + \exp(z^{i}_{NoRel})} 
\end{equation}

\tit{Multi-Consist Filter (MCT)}
After retrieving the $D_{k_{SRT}}$ documents using $[Rel]$ token, it is important to note that this token assesses the relevance of each document individually with respect to the input. However, This per-document evaluation may lead to issues such as a lack of global context, or a collection of documents that, despite being individually relevant, are not collectively coherent. To address these shortcomings, we introduce a consistency check that examines $D_{k_{SRT}}$ from a holistic perspective to filter out inconsistencies and assemble a more reliable reference set. Specifically, we employ a fine-tuned MLLM with a dedicated prompt $P_{CST}$. This process refines the collection to a subset of $ k_{CST} $ documents, providing robust external knowledge for MLLM.
\begin{equation}
D_{k_{CST}}= \text{MLLM}_{\text{ft}}(D_{k_{SRT}},P_{CST})
\end{equation}

After obtaining the external knowledge $D_{k_{CST}}$ through the \textit{token system}, we could generate answers based on this reference knowledge, similar to the standard MMRAG approach. The introduction of the \textit{token system} significantly enhances the relevance and accuracy of retrieved knowledge, thereby enabling the MLLM to produce more precise and reliable answers. This process can be summarized as follows:
\begin{equation}
    y_{\text{ans}} = \underset{y}{\arg\max}\ \text{MLLM}(y|I,Q,D_{k_{CST}},P_{vqa}).
    \label{eq:retrieval_augmented_generation_own}
\end{equation}

\subsection{Training with Token System}
\begin{figure*}[t]
    \centering
    \includegraphics[width=\textwidth]{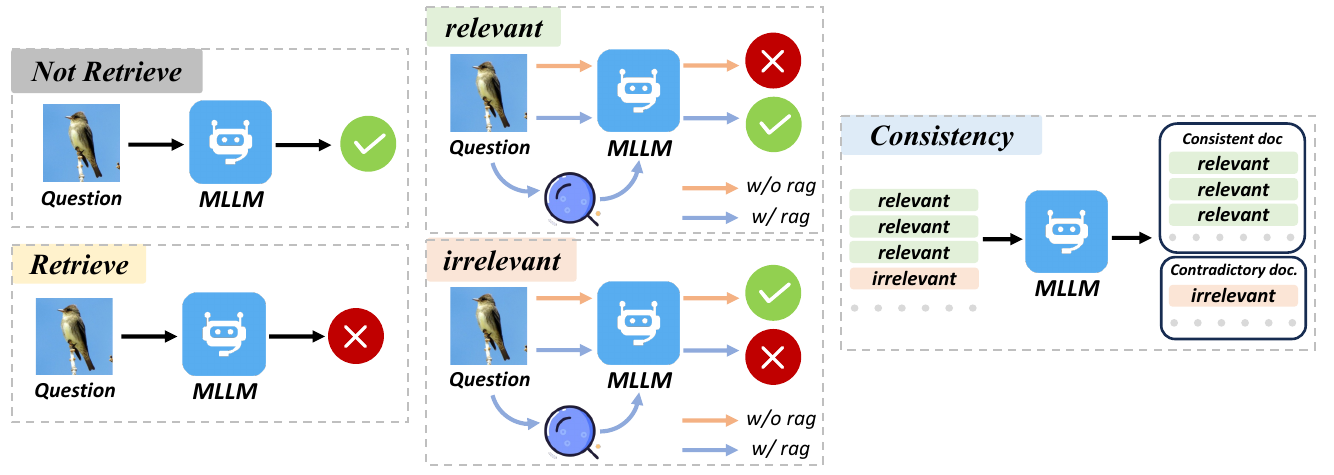}
    \vspace{-.6cm}
    \caption{Training strategy for token systems. \textbf{RET} determine whether to invoke the retrieval mechanism based on the model's response accuracy; \textbf{SRT} assess reference text quality by evaluating the model's performance when utilizing these documents as context; and \textbf{CST} uniformly evaluate all retrieved documents to deliver consistent reference documents.}
    \label{fig:model}
    \vspace{-.35cm}
\end{figure*}

In the previous section, we introduced three distinct tags to filter retrieved documents, significantly enhancing the relevance and accuracy of the reference materials and thereby improving the MLLM's response quality. Our methodology represents a significant departure from existing approaches \citep{asai2023self,cocchi2024augmenting} that rely on external annotation pipelines or auxiliary models. As illustrated in Fig.\ref{fig:model}, the principal innovation lies in our knowledge boundary-aware paradigm, where all tag determinations are explicitly derived from the target MLLM's intrinsic knowledge scope and limitations. This self-contained framework ensures that tagging decisions emerge through the model's self-assessment mechanisms rather than external supervision, thereby achieving precise alignment with the model's epistemic boundaries.

\tit{RET}
Considering the capabilities of MLLM in addressing VQA tasks, it is observed that these models inherently possess a significant amount of knowledge. This intrinsic knowledge enables them to accurately answer questions without the need for any external information retrieval. our method uniquely leverages the existing knowledge boundaries of the MLLM itself to determine whether external retrieval is required for each query. 

The detailed algorithmic process is illustrated in Algorithm \ref{alg:4}, which initializes an empty dataset and constructs a training dataset based on the model's ability to correctly respond to queries. For correctly answered queries, we label them as $[Ret]$ as the model's inherent knowledge suffices. Conversely, incorrectly answered queries are labeled as $[No Ret]$, indicating retrieval is necessary due to knowledge gaps or hallucination. Following data collection, we fine-tune the MLLM using prompt $P_{RET}$ to enable it to determine whether the target MLLM requires retrieval support for a given query.

\begin{algorithm}[t]
    \caption{Data Construction of RET}
    \label{alg:4}
    \small 
    \textbf{Input:} image-question pairs $\{(I_1, Q_1), (I_2, Q_2), \ldots, (I_N, Q_N)\}$, target MLLM \\
    \textbf{Output:} RET training dataset $D_{\text{RET}}$
    
    \begin{algorithmic}[1]
        \State Initialize an empty dataset $D_{\text{RET}} = \emptyset$
        \For{$i = 1$ to $N$}
            \State Input $(I_i, Q_i)$ into the MLLM
            \State Obtain the inference response $A_i$ from MLLM
            \If{$A_i$ is correct}
                \State Set $RET_i = [Ret]$
            \Else
                \State Set $RET_i = [No Ret]$
            \EndIf
            \State Add $(I_i, Q_i, R_i)$ to $D_{\text{RET}}$
        \EndFor
        \State \Return $D_{\text{RET}}$        
    \end{algorithmic}
    \vspace{-0.1cm}
\end{algorithm}

\tit{SRT} 
Traditional RAG systems typically employ embedding-based methods to retrieval reference documents that are semantically most relevant to the input query. However, this approach presents several limitations: (1) semantic similarity alone does not guarantee that the retrieved document will contribute to answering the question correctly, and (2) the retrieval process operates independently of the downstream MLLM, potentially introducing distractor content that adversely affects response accuracy. To overcome these limitations, we propose a novel model-centric retrieval paradigm that performs relevance assessment through the lens of the target MLLM itself, explicitly evaluating each candidate document's capacity to support accurate response generation.

As outlined in Algorithm \ref{alg:training_with_single_relevant}, we introduce a novel data collection strategy for training relevance assessment MLLM. Our method identifies relevant documents as those that convert incorrect answers to correct ones, while irrelevant documents are those that corrupt initially correct answers. This approach creates natural contrastive pairs that enable effective fine-tuning of MLLM with prompt $P_{SRT}$, optimizing their ability to discern helpful reference documents for accurate question answering.

\begin{algorithm}[t]
    \caption{Data Construction of SRT}
    \label{alg:training_with_single_relevant}
    \small 
    \textbf{Input:} image-question pairs $\{(I_1, Q_1), (I_2, Q_2), \ldots, (I_N, Q_N)\}$, target MLLM, retriever $R$ \\
    \textbf{Output:} SRT training dataset $D_{\text{SRT}}$
    \begin{algorithmic}[1] 
        \State Initialize an empty dataset $D_{\text{SRT}} = \emptyset$
        
        \For{$m = 1$ to $M$}

            \State Input $(I_m, Q_m)$ into the MLLM
            \State Obtain the inference response $A_{m,\text{direct}}$ from MLLM
                
            \State Use $R$ to retrieve top $K$ most similar documents $D_K^m = \{d_1^m, d_2^m, \ldots, d_k^m\}$ for $(I_m, Q_m)$
            \For{$k = 1$ to $K$}
                
                \State Input $(I_m, Q_m, d_k^m)$ into the MLLM
                \State Obtain the inference response $A_{m,k}$ from MLLM
                
                \If{$A_{m,\text{direct}}$ is incorrect \textbf{and} $A_{m,k}$ is correct}
                    \State Set $SRT_{m,k} = [Rel]$
                \ElsIf{$A_{m,\text{direct}}$ is correct \textbf{and} $A_{m,n}$ is incorrect}
                    \State Set $SRT_{m,k} = [NotRel]$
                \EndIf
                
                \State Add $(I_m, Q_m, d_k^m, SRT_{m,k})$ to $D_{\text{SRT}}$
            \EndFor
        \EndFor

        \State \Return $D_{\text{SRT}}$
    \end{algorithmic}
\end{algorithm}

\tit{MCT} 
Traditional methods \citep{asai2023self,cocchi2024augmenting} that rely solely on single-relevance token can only determine the relevance between a query and individual reference documents. This approach fails to address contradictions, redundancies, and inconsistencies that may arise when multiple reference documents are used simultaneously, ultimately compromising the accuracy of responses. To address this limitation, we propose the Multi-Consist mechanism, which performs unified filtering and cleaning across multiple documents, eliminating contradictions and other negative elements while preserving clean, consistent, and coherent reference content for the target MLLM, thereby enhancing response accuracy.

Algorithm \ref{alg:training_with_multi_consist} outlines our approach for constructing MCT training data from $D_{SRT}$. We first leverage the MLLM to generate concise summaries of $D_{SRT}$, effectively eliminating redundant information. To build robustness against noisy references, we intentionally contaminate $S_m^r$ by introducing $\tau\%$ of irrelevant documents $S_m^{ir}$, then identify indices of $S_m^r$ entries ranked by [Single Relevance] probability. This approach trains the model to distinguish high-quality information within mixed-quality inputs. Our objective is to enable the MLLM, when presented with the mixed-quality corpus $S_m^{mix}$, to simultaneously identify high-quality document indices and generate comprehensive summaries derived exclusively from these reliable sources.

\begin{algorithm}[t]
    \caption{Data Construction of MCT}
    \label{alg:training_with_multi_consist}
    \small
    \textbf{Input:} $D_{\text{SRT}}$,
    target MLLM, negative sample ratio $\tau\%$ \\
    \textbf{Output:} MCT training dataset $D_{MCT}$
    \begin{algorithmic}[1]
        \For{each unique $(I_m, Q_m)$ in $D_{\text{SRT}}$}
           \State Extract documents $S_m^r = \{d_k^m \mid SRT_{m,k} = [Rel]\}$ and $S_m^{ir} = \{d_k^m \mid SRT_{m,k} = [No Rel]\}$
            \If{$|S_m^{r}| > 1$} 
                \State Apply summary process: $Sum = MLLM(S_m^{r})$
                \State Merge with irrelevant: $S_m^{mix}, Idx = Mix(S_m^{r}, \tau\% S_m^{ir})$
                \State Add $(I_m, Q_m, Sum, S_m^{mix}, Idx)$ to $D_{MCT}$
            \EndIf
        \EndFor
        
        \State \Return $D_{MCT}$
    \end{algorithmic}
\end{algorithm}

\section{Experiments}
\subsection{Dataset}
\label{sec:setup}
\tit{Datasets} We evaluate our approach on four knowledge-intensive VQA benchmarks: (1) \textbf{OKVQA} \citep{marino2019ok} contains 14,000 questions across diverse knowledge categories, with 5,000 validation samples used in our experiments, evaluated using the VQA score metric \citep{agrawal2016vqavisualquestionanswering}; (2) \textbf{E-VQA} \citep{mensink2023encyclopedicvqavisualquestions} comprises 221,000 question-answer pairs linked to 16,700 fine-grained Wikipedia entities, featuring both single-hop and two-hop reasoning questions, where we use the 5,800 test samples evaluated with the BEM score \citep{bulian2022tomaytotomahtotokenlevelanswer}; (3) \textbf{InfoSeek} \citep{chen2023pretrainedvisionlanguagemodels} includes 1.3 million image-question pairs connected to approximately 11,000 Wikipedia pages, where following prior work \citep{Yan_2024}, we report results on the 73,000 validation samples using the official VQA score script; and (4) \textbf{M2KR} \citep{lin2024preflmr}, a multi-modal knowledge retrieval benchmark that processes the knowledge bases of InfoSeek and E-VQA at paragraph granularity, where we follow the established protocol using the PreFLMR retriever for comparative evaluation against the RA-VQA with PreFLMR baseline.

\tit{External Knowledge Bases} Both the InfoSeek and E-VQA datasets are supported by external knowledge bases derived from Wikipedia documents. Specifically, the E-VQA dataset is accompanied by a knowledge base consisting of 2 million Wikipedia pages. Each page includes the Wikipedia title, associated textual sections, and related images. In contrast, the InfoSeek dataset leverages a more extensive knowledge base that comprises 6 million Wikipedia entities.
In our experiments, we utilize the full 2 million-document knowledge base for E-VQA. For InfoSeek, following recent studies \cite{caffagni2024wiki,Yan_2024}, we extract a subset of 100,000 pages\footnote{The knowledge base used for InfoSeek contains the same entities as in \cite{caffagni2024wiki}.} from the original corpus of 6 million documents. This approach ensures efficient data processing while preserving the quality and coverage of the knowledge base.
For OKVQA, we employ a knowledge corpus based on Wikipedia documents selected for their pseudo-relevance as determined by M2KR \citep{lin2024preflmr}. Both the training and test passage corpora include all passages from this knowledge corpus. Evaluation of OKVQA is performed exclusively on the M2KR dataset.

\subsection{Implementation details}
\tit{Retrieved Knowledge}
Our experiments are particularly aimed at optimizing the knowledge injection and comprehension capabilities of large models, focusing on how these models process and understand external knowledge sources after retrieve.In our experimental design, we primarily utilized the EVA-CLIP-8B\citep{sun2024eva} for image retrieval.In this experiment ,we choose the  image-to-image retrieval, where we evaluate the similarity between a query image and images embedded within Wikipedia documents to retrieve the corresponding Wikipedia pages. 
To align with the ReflectiVA\citep{cocchi2024augmenting}, we set the number of retrieved web pages to 5. 
\tit{Model Architecture and Training detailed}
We employ Qwen2-VL-7B-Instruct \cite{wang2024qwen2} as our foundation model, which integrates a 675M-parameter Vision Transformer (ViT) for visual encoding with the Qwen2-7B language model for text processing. The architecture incorporates an MLP connector that inherently bridges image tokens and language representations, enabling effective multi-modal information fusion without external modules. To optimize computational efficiency, we implement LoRA \citep{hu2022lora} fine-tuning with a batch size of 512.
\tit{Training Data Collection}
In our study, we trained MMKB-RAG using subsets of the official Infoseek and E-VQA training datasets. To optimize computational efficiency, we randomly sampled 10\% of the available data for model training. Importantly, we employed the answer model itself for generating fine-tuning data without reliance on external models such as GPT-4o.

\tit{MCT type choose}
During the MCT phase of our experiments, we implemented three distinct refinement strategies: \textbf{Filter}: utilizing only documents indexed by MLLM outputs, preserving exclusively high-quality documents; \textbf{Merge}: employing only the MLLM-generated summary based on high-quality documents; and \textbf{Re-rank}: prioritizing MLLM-identified high-quality documents while retaining less consistent documents that might still contain valuable information. Comprehensive ablation studies validating the effectiveness of these strategies are presented in subsequent sections.

\begin{table*}[t]
\caption{VQA accuracy scores on E-VQA test set and InfoSeek validation set where all results from retrieval-augmented models are reported without considering any re-ranking stage to reorder retrieved web pages. \textbf{Bold} indicates state-of-the-art, \underline{underline} denotes second-best, and $\dag$ marks our reproduced results. \KB{Gray} color indicates,
results that are not directly comparable due to different knowledge bases. All retrieval-augmented results exclude re-ranking.
}
\vspace{-0.3cm}
\label{tab:results}
  \centering
  \resizebox{0.97\linewidth}{!}{
  \begin{tabular}{lc c cc c cc c ccc}
   \toprule
    & & & & & & \multicolumn{2}{c}{\textbf{E-VQA}} & & \multicolumn{3}{c}{\textbf{InfoSeek}} \\
    \cmidrule(lr){7-8} \cmidrule(lr){10-12}
     \textbf{Model} & \textbf{LLM} & & \textbf{Retriever} & \textbf{Feature} & & Single-Hop & All & & Unseen-Q & Unseen-E & All \\
    \midrule
    \multicolumn{12}{l}{\textbf{Zero-shot MLLMs}} \\
    BLIP-2~\cite{li2023blip} & Flan-T5$_\text{XL}$ & & - & - & &  12.6 & 12.4 & & 12.7 & 12.3 & 12.5 \\
    InstructBLIP~\cite{dai2023instructblip} & Flan-T5$_\text{XL}$ & & - & - & &  11.9 & 12.0 & & 8.9 & 7.4 & 8.1 \\
    LLaVA-v1.5~\cite{liu2023improved} & Vicuna-7B & & - & - & & 16.3 & 16.9 & & 9.6 & 9.4 & 9.5 \\
    LLaVA-v1.5~\cite{liu2023improved} & LLaMA-3.1-8B & & - & - & & 16.0 & 16.9 & & 8.3 & 8.9 & 7.8 \\
     Qwen2-VL-Instruct$\dag$~\cite{wang2024qwen2} & Qwen-2-7B & & - & - & & 16.4 & 16.4 & & 17.9 & 17.8 & 17.9 \\
    Qwen2-VL-Instruct(sft)$\dag$~\cite{wang2024qwen2} & Qwen-2-7B & & - & - & & 25.0 & 23.8 & & 22.7 & 20.6 & 21.6 \\
    \midrule
    \multicolumn{12}{l}{\textbf{Retrieval-Augmented Models}} \\
    Wiki-LLaVA~\cite{caffagni2024wiki} & Vicuna-7B & & CLIP ViT-L/14+Contriever & Textual & & 17.7 & 20.3 & & 30.1 & 27.8 & 28.9 \\
    Wiki-LLaVA~\cite{caffagni2024wiki} & LLaMA-3.1-8B & & CLIP ViT-L/14+Contriever & Textual & & 18.3 & 19.6 & & \underline{28.6} & 25.7 & 27.1 \\
    EchoSight~\cite{Yan_2024} & Mistral-7B/LLaMA-3-8B & & EVA-CLIP-8B & Visual & & \KB{19.4} & - & & - & - & \KB{27.7} \\
    EchoSight~\cite{Yan_2024} & LLaMA-3.1-8B & & EVA-CLIP-8B & Visual & & 26.4 & 24.9 & & 18.0 & 19.8 & 18.8 \\
   ReflectiVA~\cite{cocchi2024augmenting} & LLaMA-3.1-8B & & EVA-CLIP-8B & Visual & & \underline{35.5} & \underline{35.5} & & \underline{28.6} & \underline{28.1} & \underline{28.3} \\
    Qwen2-VL-Instruct$\dag$~\cite{wang2024qwen2} & Qwen-2-7B & & EVA-CLIP-8B + Contriever & Visual+Textual & & 25.9 & 23.6 & & 21.6 & 21.3 & 21.4 \\
    Qwen2-VL-Instruct(sft)$\dag$~\cite{wang2024qwen2} & Qwen-2-7B & & EVA-CLIP-8B + Contriever & Visual+Textual & & 32.6 & 29.9 & & 23.3 & 23.8 & 23.6 \\
    \textbf{\ours (Ours)} & Qwen-2-7B & & EVA-CLIP-8B & Visual & & \textbf{39.7} & \textbf{35.9} & & \textbf{36.4} & \textbf{36.3} & \textbf{36.4} \\
  \bottomrule
  \end{tabular}
  }
  \vspace{0.2cm}
\vspace{-0.4cm}
\end{table*}
\subsection{Comparisons with SOTA}
We evaluate our MMKB-RAG method against state-of-the-art models on the E-VQA and InfoSeek benchmarks, with results presented in Table~\ref{tab:results}. The evaluation is organized into two categories: (1) Zero-shot MLLMs without knowledge injection and (2) Retrieval-Augmented Models with knowledge injection. For baseline comparisons, we include Qwen2-VL-7B-Instruct and its fine-tuned variant Qwen2-VL-7B-Instruct(SFT) trained on the target dataset in both tasks. To establish a fair comparison with Wiki-LLaVA, we employ Contriever to retrieve the top-k documents with highest similarity to the query.

MMKB-RAG demonstrates superior performance across all settings. In the zero-shot scenario, our approach outperforms models like BLIP-2 and InstructBLIP that rely solely on pre-trained knowledge without external retrieval mechanisms. For the retrieval-augmented scenario, MMKB-RAG significantly surpasses both Qwen2-VL-7B-Instruct and its fine-tuned counterpart, despite these models sharing similar architectural foundations. This performance gap highlights the effectiveness of our proposed MMKB-RAG approach in substantially improving response accuracy.

\begin{figure*}[t]
\begin{minipage}{0.325\linewidth}
\scriptsize{\textbf{question}: In which part(s) of the world does this fish live?\vspace{0.05cm}}\\
\begin{minipage}{0.443\linewidth}
\includegraphics[width=1.\linewidth]{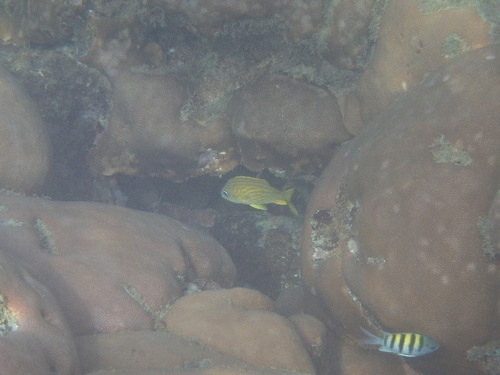}
\end{minipage}
\hfill
\begin{minipage}{0.53\linewidth}
\scriptsize{
\textbf{Qwen2-VL-Instruct}:\\
Cuba \textcolor{red}{\xmark} \\
\textbf{Qwen2-VL-Instruct(sft)}:\\
Bahía de Cochinos \textcolor{red}{\xmark} \\
\textbf{\ours (Ours):}\\
Western Atlantic \textcolor[HTML]{00b050}{\cmark}
}
\end{minipage}
\end{minipage}
\hspace{0.02cm}
\begin{minipage}{0.325\linewidth}
\scriptsize{\textbf{question}: Which architect designed this building?\vspace{0.05cm}}\\
\begin{minipage}{0.443\linewidth}
\includegraphics[width=1.\linewidth]{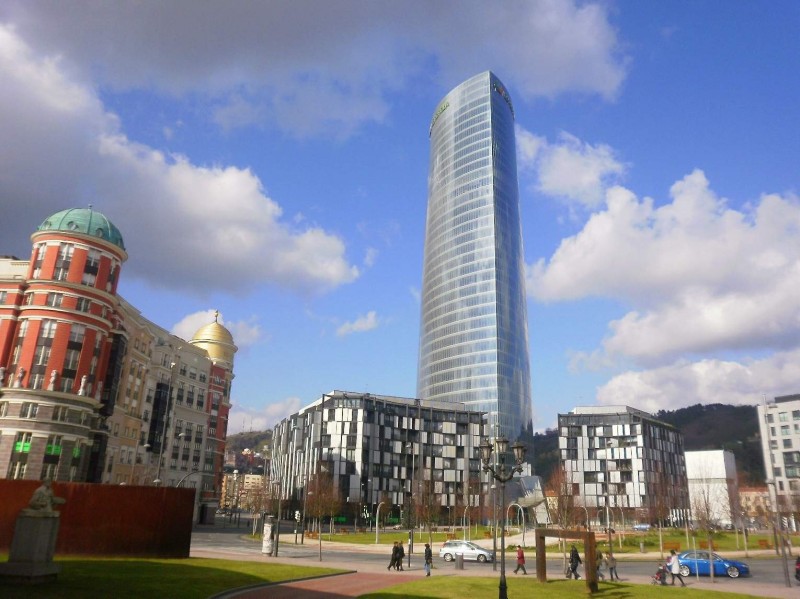}
\end{minipage}
\hfill
\begin{minipage}{0.53\linewidth}
\scriptsize{
\textbf{Qwen2-VL-Instruct}:\\
I don't know \textcolor{red}{\xmark} \\
\textbf{Qwen2-VL-Instruct(sft)}:\\
Gert Jan van Dijk \textcolor{red}{\xmark} \\
\textbf{\ours (Ours):}\\
César Pelli \textcolor[HTML]{00b050}{\cmark}
}
\end{minipage}
\end{minipage}
\hspace{0.02cm}
\begin{minipage}{0.325\linewidth}
\scriptsize{\textbf{question}: What color is the inside of this plant?\vspace{0.05cm}}\\
\begin{minipage}{0.443\linewidth}
\includegraphics[width=0.9\linewidth, height=0.8\linewidth]{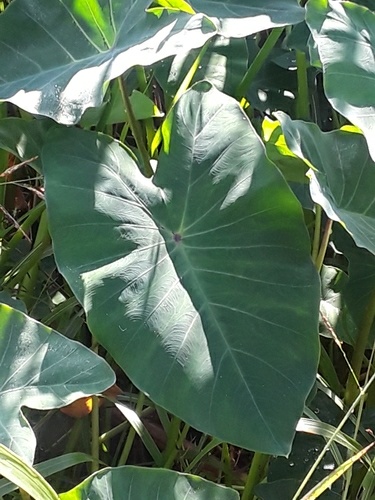}
\end{minipage}
\hfill
\begin{minipage}{0.53\linewidth}
\scriptsize{
\textbf{Qwen2-VL-Instruct}:\\
white \textcolor{red}{\xmark} \\
\textbf{Qwen2-VL-Instruct(sft)}:\\
white \textcolor{red}{\xmark} \\
\textbf{\ours (Ours):}\\
purplish \textcolor[HTML]{00b050}{\cmark}
}
\end{minipage}
\end{minipage}
\vspace{0.1cm}

\begin{minipage}{0.325\linewidth}
\scriptsize{\textbf{question}: How did this bird fare against the australian white ibis?\vspace{0.05cm}}\\
\begin{minipage}{0.443\linewidth}
\includegraphics[width=1.\linewidth, height=0.8\linewidth]{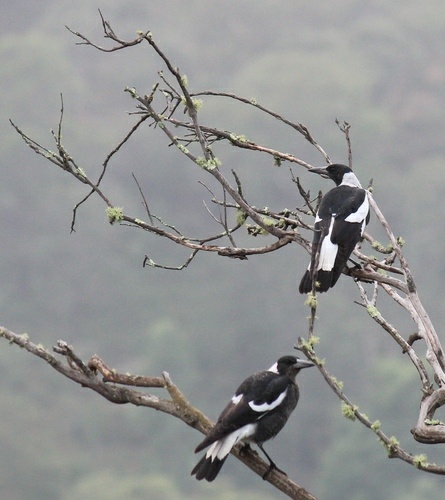}
\end{minipage}
\hfill
\begin{minipage}{0.53\linewidth}
\scriptsize{
\textbf{Qwen2-VL-Instruct}:\\
The Australian magpie has similar plumage but has red eyes and is found mainly on the ground. \textcolor{red}{\xmark} \\
\textbf{Qwen2-VL-Instruct(sft)}:\\
well \textcolor{red}{\xmark} \\
\textbf{\ours (Ours):}\\
narrowly ahead \textcolor[HTML]{00b050}{\cmark}
}
\end{minipage}
\end{minipage}
\hspace{0.02cm}
\begin{minipage}{0.325\linewidth}
\scriptsize{\textbf{question}: On what street is this building located?\vspace{0.05cm}}\\
\begin{minipage}{0.443\linewidth}
\includegraphics[width=1.\linewidth, height=0.8\linewidth]{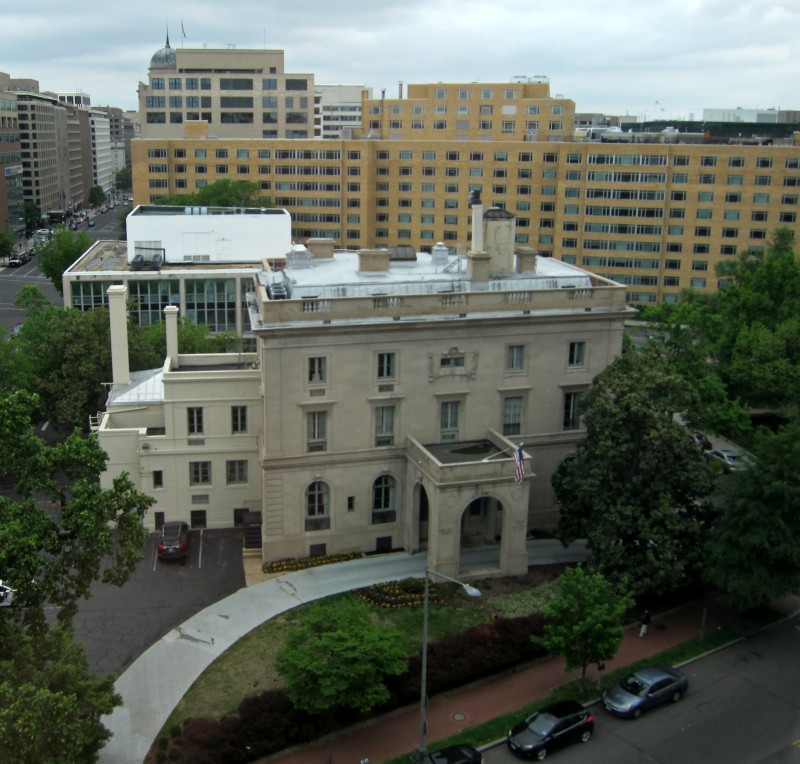}
\end{minipage}
\hfill
\begin{minipage}{0.53\linewidth}
\scriptsize{
\textbf{Qwen2-VL-Instruct}:\\
M Street \textcolor{red}{\xmark} \\
\textbf{Qwen2-VL-Instruct(sft)}:\\
M Street \textcolor{red}{\xmark} \\
\textbf{\ours (Ours):}\\
Rhode Island Avenue \textcolor[HTML]{00b050}{\cmark}
}
\end{minipage}
\end{minipage}
\hspace{0.02cm}
\begin{minipage}{0.325\linewidth}
\scriptsize{\textbf{question}: Where did former residents of this district relocate to?\vspace{0.05cm}}\\
\begin{minipage}{0.443\linewidth}
\includegraphics[width=0.9\linewidth]{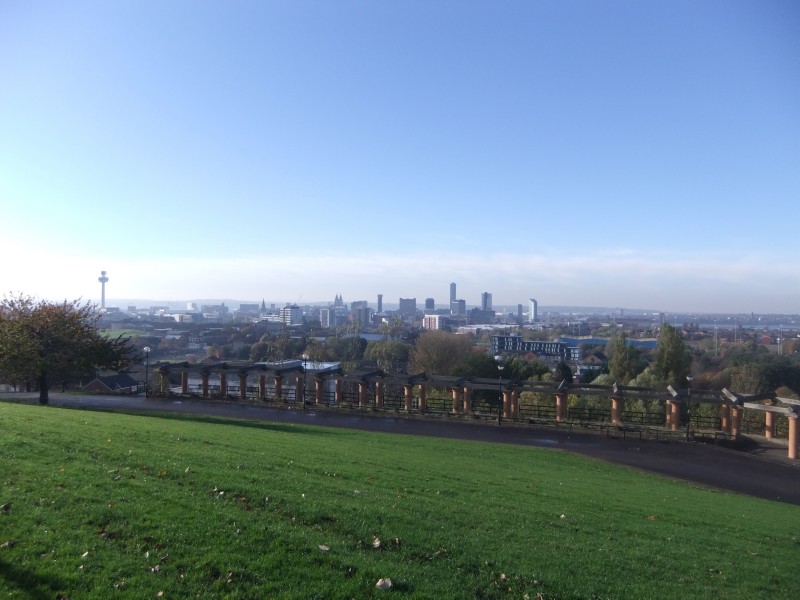}
\end{minipage}
\hfill
\begin{minipage}{0.53\linewidth}
\scriptsize{
\textbf{Qwen2-VL-Instruct}:\\
Liverpool \textcolor{red}{\xmark} \\
\textbf{Qwen2-VL-Instruct(sft)}:\\
central Liverpool \textcolor{red}{\xmark} \\
\textbf{\ours (Ours):}\\
Kirkby, Cantril Farm, and Netherley \textcolor[HTML]{00b050}{\cmark}
}
\end{minipage}
\end{minipage}
\vspace{-0.25cm}
\caption{Sample qualitative results between MMKB-RAG and w/o MMKB-RAG.}
\label{fig:qualitatives}
\vspace{0.2cm}
\end{figure*}

When utilizing the EVA-CLIP retriever for image-to-image retrieval, MMKB-RAG achieves even more substantial improvements over competitive methods such as EchoSight and Reflective that employ similar retrieval mechanisms. These results demonstrate that our method can achieve superior performance even under similar token system. Fig.\ref{fig:qualitatives} provides a qualitative comparison on different models.

\subsection{Ablation Studies and Analyses}
\subsubsection{Effectiveness of MMKB-RAG Tokens.}

\begin{figure*}[t]
    \centering
    \includegraphics[width=\textwidth]{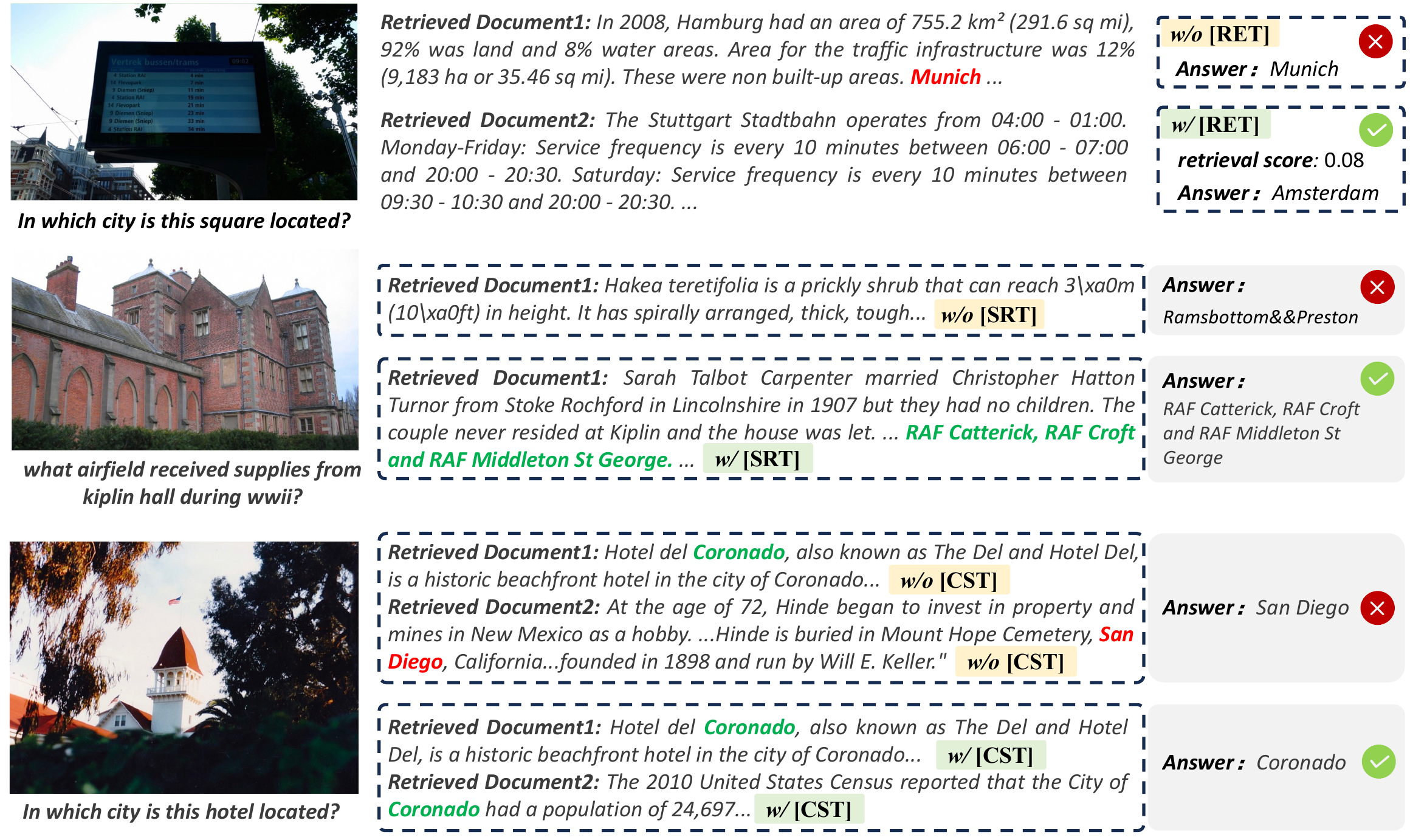}
    \vspace{-.6cm}
    \caption{Illustration of our Token System, showcasing the effects of RET, SRT, and CST. Dashed lines indicate the operational positions of different tokens, with red text highlighting erroneous information and green text denoting correct information.}
    \label{fig:ablation_visualization}
    \vspace{-.35cm}
\end{figure*}

To evaluate each token's contribution, we conducted an ablation study by progressively enabling different components of our framework. Table~\ref{tab:comparison1} summarizes these results, where $k$ represents the number of retrieved documents. When only RET is enabled, the MLLM uses the default setting ($k=5$). In contrast, when SRT is utilized, paragraphs with relevance scores above 0.5 are retained and ranked in descending order of relevance ($k=\text{auto}$).

Our experiments reveal distinctive contributions from each token: (1) \textbf{RET} improves performance across all metrics, though its impact diminishes when other tokens are introduced; (2) \textbf{SRT} significantly enhances all evaluation metrics, indicating that while embedding-based retrieval can identify semantically related documents, it cannot guarantee the accuracy of the final answers; (3) \textbf{MCT} further refines results by eliminating inconsistent or redundant documents, substantially improving performance beyond SRT alone. 

Optimal performance is achieved when all three tokens work in concert. This confirms the synergistic design of MMKB-RAG: RET provides foundational knowledge, SRT ensures relevance, and consistency filtering removes noise, collectively delivering state-of-the-art performance across both datasets. Fig.\ref{fig:ablation_visualization} provides a visualization of how each token contributes to the performance.

\begin{table}[t]
  \caption{Ablation study on the token system.}
  \vspace{-0.3cm}
  \label{tab:comparison1}
  \centering
  \setlength{\tabcolsep}{.2em} 
  \resizebox{\linewidth}{!}{ 
  \begin{tabular}{cccc cccc cc}
   \toprule
    \textbf{RET} & \textbf{SRT} & \textbf{MCT} & $k$ & \multicolumn{2}{c}{\textbf{E-VQA}} & \multicolumn{3}{c}{\textbf{InfoSeek}} \\
    \cmidrule(lr){5-6} \cmidrule(lr){7-9} 
     &  &  &  & Single-Hop & All & Unseen-Q & Unseen-E & All \\
    \midrule
     &  &  & 5 & 32.57 & 29.88 & 23.33 & 23.82 & 23.57 \\
    \checkmark &  &  & 5 & 32.72 & 30.01 & 24.55 & 24.59 & 24.57 \\
     & \checkmark &  & auto & 38.09 & 34.63 & 34.07 & 33.55 & 33.81 \\
    \checkmark & \checkmark &  & auto & 38.13 & 34.66 & 34.07 & 33.56 & 33.81 \\
     & \checkmark & \checkmark & auto & 39.62 & 35.89 & 36.43 & 36.32 & 36.37 \\
     \checkmark & \checkmark & \checkmark & auto & \textbf{39.66} & \textbf{35.91} & \textbf{36.44} & \textbf{36.34} & \textbf{36.37} \\
    \bottomrule
  \end{tabular}
  }
  \vspace{-0.4cm}
\end{table}
\subsubsection{Effectiveness of document quantity}

In this section, we investigate the impact of document numbers $k$ on the performance of SRT and MCT strategies. To isolate the effect of $k$, we fix the retrieval module to be always enabled. The results are presented in Tables~\ref{tab:single_relevant} and~\ref{tab:multi_consist}.

\tit{SRT} For E-VQA, $k=5$ achieves peak performance, suggesting that moderate document quantities effectively balance information completeness and noise reduction. The InfoSeek dataset similarly favors $k=5$, though performance degrades sharply at $k=20$, indicating that the automatic selection mechanism may inadvertently discard critical information.

\tit{MCT} On InfoSeek, merge-based MCT demonstrates superior performance as it emphasizes precise answers by more accurately identifying consistent documents and eliminating redundant information. In contrast, rerank-based MCT excels on E-VQA under BEM metrics, which emphasize semantic similarity rather than exact matching. The rerank strategy proves advantageous in this context as it prioritizes documents demonstrating higher consistency while preserving information breadth, thus minimizing the risk of omitting critical details that might occur with other approaches.
 
\subsubsection{Effectiveness of Training Data Construction}

To evaluate the impact of different training data construction methods, we compare our approach with a more powerful external model (GPT-4o) on the SRT module, which has the most significant influence on system performance. Due to cost constraints, we limited GPT-4o-generated samples to 2,000, while our method produced datasets ranging from 2,000 to 100,000 samples. Table~\ref{tab:gpt4} summarizes the results.

Our experiments reveal two key findings: First, increasing training data volume with our method does not yield performance improvements. This suggests that small, high-quality datasets are sufficient for the model to learn effective preference patterns. Second, with equal training data volume (2,000 samples), our method significantly outperforms GPT-4o despite the latter's superior general capabilities. We hypothesize that our construction approach produces training data that better aligns with the answering model's characteristics, resulting in more effective document re-ranking and scoring.

\begin{table}[t]
\caption{Impact of varying the number of retrieved documents on SRT performance.}
\vspace{-0.3cm}
  \label{tab:single_relevant}
  \centering
  \resizebox{\linewidth}{!}{
  \tiny
  \begin{tabular}{c ccc ccc}
   \toprule
    $k$ & \multicolumn{2}{c}{\textbf{E-VQA}} & \multicolumn{3}{c}{\selectfont\textbf{InfoSeek}} \\
    \cmidrule(lr){2-3} \cmidrule(lr){4-6}
     &  Single-Hop &  All &  Unseen-Q &  Unseen-E &  All \\
    \midrule
     auto & 38.09 & 34.63 & 34.07 & 33.55 & 33.81\\
     1 & 37.30 & 33.79 & 33.39 & 30.77 & 32.03 \\
     5 & \textbf{39.56} & \textbf{35.89} & \textbf{35.05} & \textbf{34.91} & \textbf{34.98} \\
     10 & 38.90 & 35.33 & 33.68 & 33.18 & 33.43 \\
     15 & 37.39 & 33.84 & 30.16 & 29.82 & 29.99 \\
     20 & 35.20 & 32.08 & 25.32 & 25.09 & 25.09 \\
    \bottomrule
  \end{tabular}
  }
   \vspace{-0.4cm}
\end{table}

\begin{table}[t]
\caption{Impact of Different MCT Types on Performance.}
\vspace{-0.3cm}
  \label{tab:multi_consist}
  \centering
  \resizebox{\linewidth}{!}{ 
  \begin{tabular}{c c ccc ccc}
   \toprule
    \textbf{MCT type} & $k$ & \multicolumn{2}{c}{\textbf{E-VQA}} & \multicolumn{3}{c}{\textbf{InfoSeek}} \\
    \cmidrule(lr){3-4} \cmidrule(lr){5-7} 
     &  & Single-Hop & All & Unseen-Q & Unseen-E & All \\
     \midrule
    merge & auto & 34.84 & 32.14 & \textbf{36.43} & \textbf{36.32} & \textbf{36.37} \\
    rerank & auto & \textbf{39.62} & \textbf{35.90} & 34.53 & 34.96 & 34.74 \\
    filter & auto & 39.47 & 35.74 & 35.82 & 35.04 & 35.43 \\
    \midrule
    merge & 5 & 35.77 & 33.00 & \textbf{36.10} & \textbf{35.80} & \textbf{36.00} \\
    rerank & 5 & \textbf{39.16} & \textbf{35.43} & 35.17 & 34.98 & 35.07 \\
    filter & 5 & 38.48 & 35.00 & 34.81 & 34.08 & 34.44 \\
     \midrule
    merge & 10 & 34.61 & 31.79 & \textbf{34.38} & \textbf{34.48} & \textbf{34.43} \\
    rerank & 10 & 38.61 & 35.08 & 32.83 & 32.96 & 32.89 \\
    filter & 10 & \textbf{38.78} & \textbf{35.20} & 33.52 & 32.76 & 33.14 \\
     \midrule
    merge & 15 & 33.47 & 30.57 & \textbf{35.91} & \textbf{35.65} & \textbf{35.78} \\
    rerank & 15 & 37.16 & 33.72 & 30.09 & 29.90 & 29.99 \\
    filter & 15 & \textbf{37.81} & \textbf{34.52} & 35.11 & 33.99 & 34.54 \\
     \midrule
    merge & 20 & 31.98 & 29.64 & \textbf{35.61} & \textbf{35.04} & \textbf{35.32} \\
    rerank & 20 & \textbf{35.32} & 32.14 & 25.20 & 25.17 & 25.18 \\
    filter & 20 & 35.24 & \textbf{32.35} & 34.72 & 33.33 & 34.01 \\
    \bottomrule
  \end{tabular}
  }
  \vspace{-0.3cm}
\end{table}

\begin{table}[t]
  \caption{Comparison of SRT's performance trained with GPT-4o and our proposed method.}
  \vspace{-0.3cm}
  \label{tab:gpt4}
  \centering
  \resizebox{\linewidth}{!}{
  \begin{tabular}{lccccccc} 
   \toprule
    &  &  & \multicolumn{2}{c}{\textbf{E-VQA}} & \multicolumn{3}{c}{\textbf{InfoSeek}} \\
    \cmidrule(lr){4-5} \cmidrule(l){6-8}
    & num & $k$ & Single-Hop & All & Unseen-Q & Unseen-E & All \\
    \midrule
    GPT-4o & 2k & auto & 39.81 & 36.13 & 33.03 & 33.22 & 33.12 \\
    
    \midrule
    
    \multirow{5}{*}{\textbf{MMKB-RAG}} & 2k & auto & \textbf{40.06} & \textbf{36.24} & \textbf{35.22} & \textbf{36.65} & \textbf{35.92} \\
     & 5k & auto & 39.68 & 35.68 & 34.78 & 34.72 & 34.75 \\
     & 10k & auto & 39.22 & 35.50 & 33.73 & 33.93 & 33.83 \\
     & 50k & auto & 39.24 & 35.39 & 34.85 & 34.14 & 34.49 \\
     & 100k & auto & 38.09 & 34.63 & 34.07 & 33.55 & 33.81 \\
    \bottomrule
  \end{tabular}
  }
  \vspace{-0.3cm}
\end{table}

\subsection{Comparison on M2KR dataset}
We further validate our approach on the M2KR dataset, which is specifically designed to benchmark multimodal models on image+text-to-text retrieval tasks. In this evaluation, models must generate responses by jointly reasoning over visual and textual inputs. Table~\ref{tab:model_comparison_scores} presents a comparative analysis of our MMKB-RAG method against strong baselines, including RA-VQAv2 and various Qwen2-VL-Instruct variants, all leveraging the PreFLMR retriever. The results demonstrate that MMKB-RAG consistently outperforms both RA-VQAv2 and the baseline knowledge-based models across evaluation metrics, validating the effectiveness of our approach.

\begin{table}[t]
 \caption{Performance comparison on the M2KR dataset.}
 \vspace{-0.3cm}
  \label{tab:model_comparison_scores}
  \centering
  \small
  \begin{tabular}{lccc}
   \toprule
    \textbf{Model} & \textbf{OKVQA} & \textbf{Infoseek} & \textbf{E-VQA} \\
    \midrule
    \multicolumn{4}{l}{\textbf{Zero-shot MLLMs}} \\ 
    RA-VQAv2  & 55.44 & 21.78 & 19.80 \\
    Qwen2-VL-Instruct  & 60.45 & 21.75 & 19.01 \\
    Qwen2-VL-Instruct(sft)  & 64.08 & 26.00 &  26.72 \\
    \midrule
    \multicolumn{4}{l}{\textbf{Retrieval-Augmented Models}} \\ 
    RA-VQAv2 w/ FLMR & 60.75 & - & - \\
    RA-VQAv2 w/ PreFLMR & 61.88 & 30.65 & 54.45 \\
    Qwen2-VL-Instruct w/ PreFLMR  & 46.99 & 24.68 & 51.81 \\
    Qwen2-VL-Instruct(sft) w/ PreFLMR  & 65.07 & 30.74 & 53.89 \\
    \textbf{MMKB-RAG w/ PreFLMR} & \textbf{65.44} & \textbf{34.72} & \textbf{60.93} \\
    \bottomrule
  \end{tabular}
  \vspace{-0.4cm}
\end{table}

\section{Conclusion}
In this study, we introduce a new framework called multi-modal Knowledge-Based Retrieval-Enhanced Generation (MMKB-RAG). This framework leverages the knowledge boundary of the answer model to dynamically generate tags for the RAG system, which enables more efficient filtering of retrieved documents and retaining only the most relevant and accurate references. By doing so, MMKB-RAG significantly improves the accuracy and robustness of model responses in multi-modal tasks.


\bibliographystyle{ACM-Reference-Format}
\bibliography{sample-sigconf-authordraft}

\appendix
\end{document}